\documentclass[]{spie}  %>>> use for US letter paper
\usepackage[]{graphicx}
\usepackage[]{amsmath}
\usepackage{multicol}
\usepackage{float}
\usepackage{hyperref}
\usepackage{authblk}

\begin{document}

\title{Radon Inversion via Deep Learning}

\author[a,b]{Ji~He}
\author[a,b]{Jianhua~Ma}
\affil[a]{Department of Biomedical Engineering, Southern Medical University, Guangzhou, 510515, China}
\affil[b]{Guangzhou Key Laboratory of Medical Radiation Imaging and Detection Technology, Guangzhou, 510515, China}

%>>>> Further information about the authors, other than their
%  institution and addresses, should be included as a footnote,
%  which is facilitated by the \authorinfo{} command.

\authorinfo{Correspondence: J.Ma: E-mail: jhma@smu.edu.cn. This work was supported in part by the National Natural Science Foundation of China under Grant Nos. U1708261 and 61571214.}
%%%%%%%%%%%%%%%%%%%%%%%%%%%%%%%%%%%%%%%%%%%%%%%%%%%%%%%%%%%%%

%%>>>> when using amstex, you need to use @@ instead of @

%%%%%%%%%%%%%%%%%%%%%%%%%%%%%%%%%%%%%%%%%%%%%%%%%%%%%%%%%%%%%
%>>>> uncomment following for page numbers
% \pagestyle{plain}
%>>>> uncomment following to start page numbering at 301
%\setcounter{page}{301}

\maketitle

%%%%%%%%%%%%%%%%%%%%%%%%%%%%%%%%%%%%%%%%%%%%%%%%%%%%%%%%%%%%%
\begin{abstract}
Radon transform is widely used in physical and life sciences and one of its major applications is the X-ray computed tomography (X-ray CT), which is significant in modern health examination. The Radon inversion or image reconstruction is challenging due to the potentially defective radon projections. Conventionally, the reconstruction process contains several \emph{ad hoc} stages to approximate the corresponding Radon inversion. Each of the stages is highly dependent on the results of the previous stage. In this paper, we propose a novel unified framework for Radon inversion via deep learning (DL). The Radon inversion can be approximated by the proposed framework with an end-to-end fashion instead of processing step-by-step with multiple stages. For simplicity, the proposed framework is short as iRadonMap (\textbf{i}nverse \textbf{Radon} transfor\textbf{m} \textbf{ap}proximation). Specifically, we implement the iRadonMap as an appropriative neural network, of which the architecture can be divided into two segments. In the first segment, a learnable fully-connected filtering layer is used to filter the radon projections along the view-angle direction, which is followed by a learnable sinusoidal back-projection layer to transfer the filtered radon projections into an image. The second segment is a common neural network architecture to further improve the reconstruction performance in the image domain. The iRadonMap is overall optimized by training a large number of generic images from ImageNet database. To evaluate the performance of the iRadonMap, clinical patient data is used. Qualitative results show promising reconstruction performance of the iRadonMap.
\end{abstract}


\section{Introduction}
In physical and life sciences, the reconstruction problem, which is to determine the internal structure or some property of the internal structure of an object without having to macroscopically damage the object, is essential. In 1917, Johann Radon of Austria presented a solution to the reconstruction problem with Radon transform and the corresponding inversion formula\cite{radon20051}. Up to now, Radon transform has been used in various applications, including the removal of multiple reflections\cite{thorson1985velocity}\cite{sacchi1995high}, regional and global seismology\cite{gorman1999wave}\cite{gu2009mantle}, astrophysics\cite{bracewell1956strip}, and computed tomography (CT)\cite{cormack1963representation}. Among these applications, X-ray CT is one of the most important branches of Radon transform. The X-ray CT, which has great advantages in various pathological diagnoses, is an indispensible imaging modality in modern hospitals and clinics.

Radon inversion or the so-called image reconstruction in X-ray CT is challenging because the radon projections acquired with physical sensors or detectors are probably defective and with noise. Conventionally, the reconstruction process contains multiple \emph{ad hoc} stages to approximate the corresponding Radon inversion. Each of these stages highly depends on the processing results of its previous stage. The processing chain of X-ray CT reconstruction includes the logarithm transformation, scatter correction, beam hardening correction, partial volume effect correction, image reconstruction, and image postprocessing. It is expected that the reconstruction performance will be largely affected if any one of these stages suffer from slight compromise.

In order to obtain promising reconstruction performance for X-ray CT, various advanced algorithms separately addressing the multiple \emph{ad hoc} stages have been proposed. Among these stages, the image reconstruction is one of the most vigour research fields. Regarding the image reconstruction, the most popular method in commercial X-ray CT scanner is filtered back-projection (FBP) algorithm. With a simple filtering operation and back-projection, the FBP algorithm can fleetly reconstruct CT images. The FBP algorithm can achieve promising reconstruction performance if high-quality radon projections could be obtained with adequate X-ray radiation exposure. However, in low-dose X-ray CT (LdCT) imaging, the reconstruction results of FBP algorithm will suffer from severe noise-induced artifacts. To obtain promising reconstruction performance, one can design more elaborate filtering operations for FBP algorithm, which is often not an easy task. In addition, the off-the-shelf denoising algorithms can be adopted to postprocess the reconstructed CT images. These denoising algorithms can improve the image quality to some extent, but the severe streak-like artifacts might be preserved as intrinsic textures.

Currently, model-based iterative reconstruction (MBIR) algorithms are widely studied for LdCT imaging. Usually, the MBIR algorithms simultaneously model both the noise properties of the radon projections and some prior knowledge of the radon projections and/or objective image, which results in a cost function comprised of two terms, namely, a data-fidelity term and a penalty or prior term. Most of the MBIR algorithms assume that two sources of noise, namely, the intrinsic quanta noise and electronic noise, are responsible to the degradation of the radon projections. This results in a similar data-fidelity term for most of the MBIR algorithms. By designing different prior knowledge for the penalty term, these MBIR algorithms can obtain promising reconstruction performances to different extent. Nevertheless, three drawbacks are accompanied with these MBIR algorithms. First, proper distributions have to be assumed to model the noise in the radon projections. Currently, the Poisson and Gaussian distributions are widely used for the intrinsic quanta noise and electronic noise, respectively. However, the real noise distribution is much more complicate than the simple Poisson and Gaussian distributions. Second, in order to obtain promising reconstruction performance, proper prior knowledge has to be designed to constrain solution space of the MBIR algorithms, which is often nontrivial. Third, these MBIR algorithms often involve several projections and back-projections during the optimizations, which will be more time-consuming than the FBP algorithm.

In addition to the respective drawbacks of the FBP and MBIR algorithms, the commonality of these two kinds of algorithms is that they both rely on a priorly calculated projection operator, which largely determines the precision of the final reconstruction. Recently, data-driven image reconstruction is receiving more attention. In \cite{zhu2018image}, Zhu \emph{et al}. proposed a data-driven supervised learning for image reconstruction, which is named as automated transform by manifold approximation (AUTOMAP). Without incorporating a priorly calculated projection operator, the AUTOMAP directly learns a mapping between the sensor and the image domain, which is emerged from an appropriate corpus of training data. Compared to FBP and MBIR algorithms, the AUTOMAP is a unified framework for image reconstruction, which can simultaneously consider the separate \emph{ad hoc} stages in image reconstruction to optimize the final reconstruction performance. However, the AUTOMAP might be unsuitable in clinical CT applications with normal dimension, i.e., $512\times 512$, because of inherence defect of the network architecture of AUTOMAP. The fully-connected layers adopted in the AUTOMAP require a huge amount of computing resource in order to train a model with reasonable reconstruction performance.

Inspired by the concept of unified image reconstruction of the AUTOMAP, in this work we propose a unified framework for Radon inversion that overcomes the drawbacks of the AUTOMAP in reconstructing images from radon projections with large dimensions. For simplicity, the proposed framework is short as RAINAP (Radon inversion approximation). The RAINAP is implemented with an appropriative neural network that based on the corresponding Radon transform. Specifically, the RAINAP can be divided into two segments. In the first segment, we parameterize the filtering operation and back-projection of the FBP algorithm with two learnable appropriative network layers to perform the domain transform, i.e., from radon projections to image domain. In the second segment, we use a convolutional neural network (CNN) to further refine the reconstruction performance. The two segments are overall optimized using a large number of training data to guarantee a promising reconstruction performance.

%%%%%%%%%%%%%%%%%%%%%%%%%%%%%%%%%%%%%%%%%%%%%%%%%%%%%%%%%%%%%
\section{METHODOLOGY}
\subsection{Theoretical Foundation}
In this section, we present the theoretical foundation for the network architecture of the iRadonMap. Without loss of generality, the Radon transform discussed in this work is with a parallel-beam X-ray CT imaging geometry. The network architecture of the iRadonMap for X-ray CT imaging geometries of fan-beam and cone-beam can be deduced similarly.

For an arbitrary 2D object $f(x,y)$, the corresponding Radon transform with a parallel-beam X-ray CT imaging geometry can be written as follows:
\begin{equation}
\small
\label{Eq_1}
p(s,\theta)=\int\limits_{-\infty}^{\infty}\int\limits_{-\infty}^{\infty}{f(x,y)\delta(xcos\theta+ysin\theta-s)dxdy}.
\end{equation}
\noindent Here, $p(s,\theta)$ denotes a radon projection of $f(x,y)$ at a certain view-angle $\theta$. $\delta(\cdot)$ is a Dirac function. $s$ is the position of a detector unit relative to the geometry center of the X-ray imaging system. Thus, $f(x,y)\delta(xcos\theta+ysin\theta-s)$ presents the intersection of an X-ray beam with $f(x, y)$. The radon projections are usually collected within rotation interval of 180 degree, namely, $0\le\theta\le \pi$.

To reconstruct $f(x, y)$, one can use FBP algorithm. Let's denote the two-dimensional Fourier transform of $f(x,y)$ as $F(\omega,\theta)$ and the one-dimensional Fourier transform of $p(s,\theta)$ as $P(\omega,\theta)$. According to Inverse Fourier transform and Central slice theorem, $f(x,y)$ can be expressed as follows:
\begin{equation}
\small
\label{Eq_2}
\begin{array}{l}
f(x,y)=\int\limits_{0}^{2\pi}\int\limits_{0}^{\infty}{F(\omega,\theta)e^{2{\pi}i\omega(xcos\theta+ysin\theta)}{\omega}d{\omega}d\theta}
\\[4mm]
\qquad\quad\,=\int\limits_{0}^{\pi}\int\limits_{-\infty}^{\infty}{P(\omega,\theta)|{\omega}|e^{2{\pi}i\omega(xcos\theta+ysin\theta)}d{\omega}d\theta}.
\end{array}
\end{equation}
\noindent Here, $|{\omega}|$ is the transfer function of the ramp filter. Introducing $Q(\omega,\theta)=|{\omega}|P(\omega,\theta)$ and denoting the Inverse Fourier transform of $Q(\omega,\theta)$ as $q(s, \theta)$, the reconstruction of an arbitrary 2D object $f(x,y)$ via FBP algorithm can be obtained with the following two steps:

1) apply a ramp filtering operation to $p(s,\theta)$ with respect to the variable $s$ in the Fourier domain, as follows:
\begin{equation}
\small
\label{Eq_3}
q(s, \theta)=FT^{-1}\{|{\omega}|{\cdot}FT\{p(s,\theta)\}\};
\end{equation}
\indent 2) back-project $q(s, \theta)$ to obtain the reconstruction, as follows:
\begin{equation}
\small
\label{Eq_4}
f(x,y)=\int\limits_{0}^{\pi}{q(s, \theta)|_{s=xcos\theta+ysin\theta}d\theta}.
\end{equation}
Here, $s=xcos\theta+ysin\theta$ denotes a sinusoidal track, from which the radon projection points are related to the reconstructed point $(x,y)$. The reconstruction by the FBP algorithm might suffer from noise-induced artifacts due to the degradation of the radon projections. In order to obtain a promising reconstruction, one can apply some off-the-shelf restoration algorithms in the radon projections and/or image domain to further improve the FBP results. This indicates that the Radon inversion can be approximated by several successive operations, each of which is highly dependent on the results of the previous operation. Recently, Zhu \emph{et al}. proposed a unified framework for image reconstruction, namely, AUTOMAP\cite{zhu2018image}, which is also suitable for Radon inversion. With it, one can reconstruct an image from the radon projections in an end-to-end fashion instead of step-by-step with multiple stages. However, it can be difficult to implement for large-size CT images (e.g., $512\times 512$), due to the stack of fully-connected layers in the AUTOMAP, which might cost huge amount of storages and computations. In order to address this issue, in this work we propose a novel unified framework (iRadonMap) to approximate the Radon inversion of large-size CT image with far less parameters than the AUTOMAP.

\begin{figure*}
  \centering
  % Requires \usepackage{graphicx}
  \includegraphics[scale=1.5]{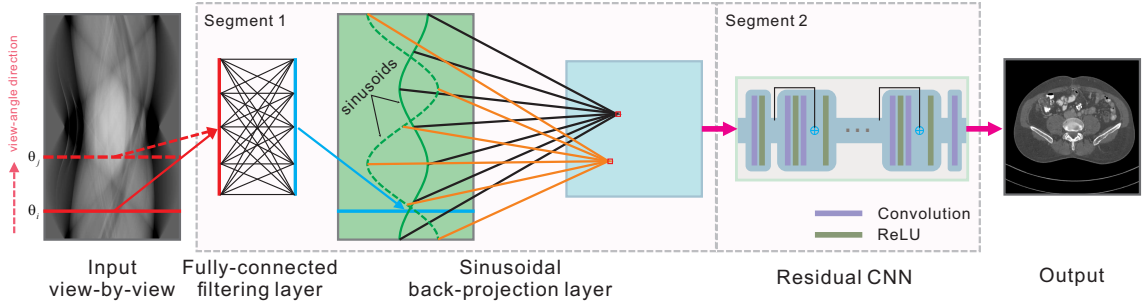}\\
  \caption{Network architecture of the iRadonMap.}\label{Fig_1}
\end{figure*}

\subsection{Network Architecture}
As shown in Fig. \ref{Fig_1}, the proposed iRadonMap can be divided into two segments. In the first segment, we use two learnable appropriative network layers to simulate the filtering operation and back-projection in FBP algorithm, respectively. The ramp filtering operation in the FBP algorithm is applied to the radon projections along the view-angle direction, as shown in Eq. (\ref{Eq_3}). In the proposed iRadonMap, we use a learnable fully-connected layer to simulate the filtering operation, which is also applied to the radon projections along the view-angle direction. Regarding the back-projection, we could simply simulate it with another fully-connected operation\cite{zhu2018image}. However, the fully-connected operation is difficult to implement for Radon inversion of large-size CT images, which involves huge number of parameters and may consume huge amount of computing resources. To address this problem, we use an appropriative network layer to simulate the back-projection. By analysing the back-projection of the FBP algorithm in Eq. (\ref{Eq_4}), it can be observed that a certain reconstruction point only relates to a few radon projected points, which constitute a sinusoid in radon projection domain. This allows us to construct an appropriative learnable operation for back-projection with far fewer parameters than the fully-connected operation. This is denoted as a sinusoidal back-projection layer.

In the second segment, we use a residual CNN to strengthen the reconstruction of the iRadonMap. The residual CNN used in the second segment of the iRadonMap is a fully convolutional network, which adopts the shortcut technique in the ResNet proposed by He \emph{et al}. \cite{He2015Deep}. Unlike the ResNet, we do not use batch normalization. It is noted that the CNN architecture used in this work for the second segment is not the only possible implementation. Other well-defined architectures, such as autoencoders, U-net can also be adopted.

\subsection{Network Training}

To obtain promising reconstruction performance, the iRadonMap is overall optimized by minimizing the mean square error (MSE), which is defined as follows:
\begin{equation}
\label{Eq_5}
E(\Theta)=\frac{1}{N}\sum\limits_{i=1}^{N}\|\bar x_{i}(\Theta)-x_{i}^{ref}\|_{2}^{2}.
\end{equation}
\noindent Here, $\bar x$ is the final output of the iRadonMap and $x^{ref}$ is the reference image. $N$ is the number of image pairs used for training. $\Theta$ represents the learnable parameters in the iRadonMap. This minimization problem can be solved with various off-the-shelf algorithms, and in this work the RMSProp algorithm (see \emph{http://www.cs.toronto.edu/$\sim$tijmen/csc321/slides/lecture\_slides}\emph{\_lec6.pdf}) is adopted. The corresponding minibatch size, learning rate, momentum, and weight decay are set to 2, 0.00002, 0.9, and 0.0, respectively. The iRadonMap is implemented on PyTorch deep learning framework \cite{paszke2017automatic}. The iRadonMap is trained for one week using two NVIDIA Tesla P40 graphics processing units (GPUs) with 24 GB memory capacity each.

\begin{figure*}
  \centering
  % Requires \usepackage{graphicx}
  \includegraphics[scale=0.7]{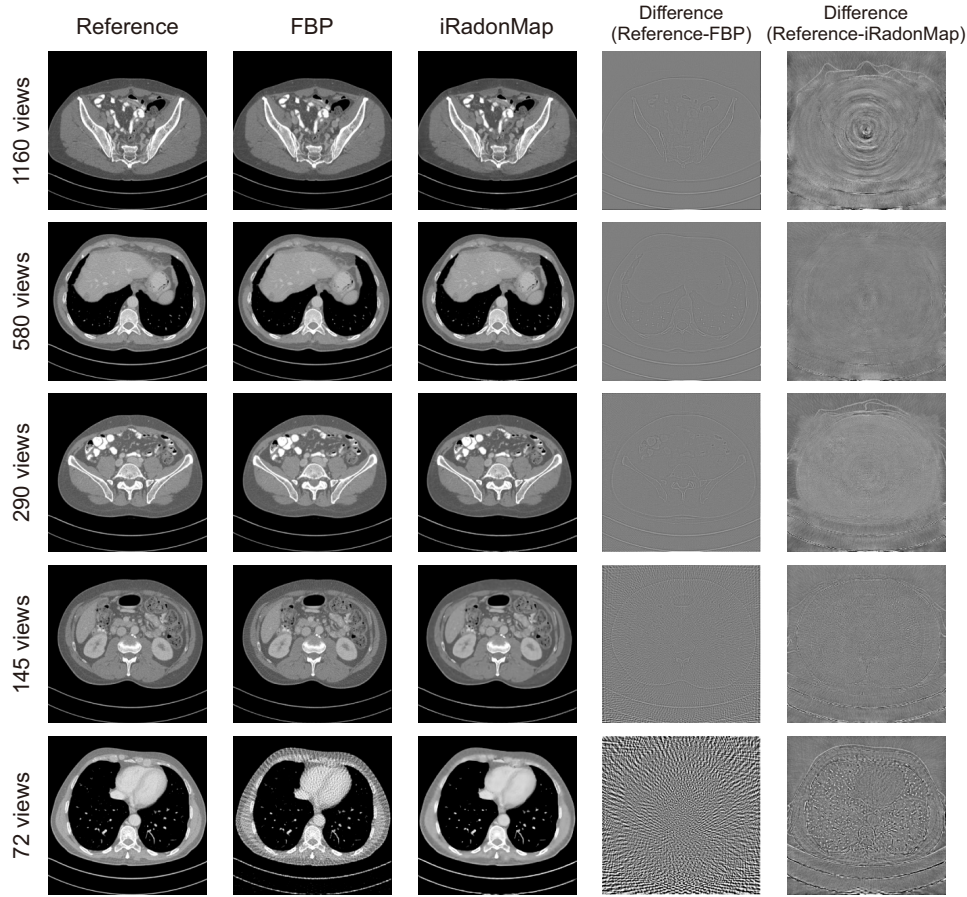}\\
  \caption{Comparison Results of Radon inversion between FBP and the proposed iRadonMap. The reconstructions are displayed with windows of [0 0.4]; the difference images are displayed with windows of [-0.08 0.08].}\label{Fig_2}
\end{figure*}

\subsection{Datasets}
The training dataset of the iRadonMap is consisted of a large number of generic images from ImageNet\cite{deng2009imagenet}. In this work, we collected 62,899 RGB color images. Similar to \cite{zhu2018image}, the Y-channel luminance of the RGB color images were extracted to form greyscale intensity images. Then, the greyscale intensity images were cropped to the central $512\times 512$ pixels. The radon projections with size of $1160\times 736$ (i.e., 1160 views with 736 detector bins each) of each image is generated with the parallel geometry in the ASTRA Toolbox\cite{Van2016Fast}. To summarize, in this work a radon projection map with size of $1160\times 736$ is feed into the iRadonMap, which outputs a reconstructed image with size of $512\times 512$. The testing dataset of the iRadonMap is consisted of real clinical patient data, which are provided and authorized by the Mayo Clinic. To demonstrate generalizability of the iRadonMap, no clinical patient data was involved in the training phase. We only test network performance with the clinical patient data.

\section{RESULTS}
In practice, different iRadonMap models should be trained for Radon transforms with different imaging geometries. In the preliminary experiments, we study the same parallel imaging geometry but with different number of view-angles, namely, 1160, 580, 290, 145, 72 views. The corresponding results are presented in Fig. \ref{Fig_2}. Specifically, the results of different views are showed in rows from up to down, respectively. The columns from left to right show the Reference, FBP, iRadonMap, different image between the Reference and FBP, and different image between the Reference and iRadonMap, respectively. By comparing the reconstruction results of the iRadonMap and FBP in row 1 to row 3, we can see that the performance of iRadonMap is similar to FBP, which indicates a robust Radon inversion performance of the proposed iRadonMap. Moreover, by comparing the different images in columns 4 and 5, it can be observed that the iRadonMap can preserve more edge details than the FBP. In row 4 and 5, it can be observed that the FBP algorithm suffers from severe streak artifacts with sparse view-angles radon projection data. On the contrary, the reconstruction performance of the iRadonMap is promising with less streak artifacts, compared to that of the FBP.

\section{CONCLUSION}
In this paper, we propose a novel framework for Radon inversion via deep learning, namely, iRadonMap. The iRadonMap is an appropriative neural network that consists of two segments. The first segment contains a learnable fully-connected filtering layer and a learnable sinusoidal back-projection layer. The second segment is a common CNN architecture. The preliminary results with clinical data show that the iRadonMap can achieve promising performance in terms of qualitative measurements. More experiments are undergoing.

%%%%%%%%%%%%%%%%%%%%%%%%%%%%%%%%%%%%%%%%%%%%%%%%%%%%%%%%%%%%%
%%%%% References %%%%%

\bibliography{report} % bibliography data in report.bib
\bibliographystyle{spiebib} % makes bibtex use spiebib.bst

\end{document}